\def\year{2021}\relax
\documentclass[letterpaper]{article} % DO NOT CHANGE THIS
\usepackage{aaai21}  % DO NOT CHANGE THIS
\usepackage{times}  % DO NOT CHANGE THIS
\usepackage{helvet} % DO NOT CHANGE THIS
\usepackage{courier}  % DO NOT CHANGE THIS
\usepackage[hyphens]{url}  % DO NOT CHANGE THIS
\usepackage{graphicx} % DO NOT CHANGE THIS
\usepackage{natbib}  % DO NOT CHANGE THIS AND DO NOT ADD ANY OPTIONS TO IT
\usepackage{caption} % DO NOT CHANGE THIS AND DO NOT ADD ANY OPTIONS TO IT

\usepackage{amsmath}
\usepackage{color}

\title{Confidence-aware Non-repetitive Multimodal Transformers for TextCaps}
\author{
    Zhaokai Wang\textsuperscript{\rm 1}, 
    Renda Bao\textsuperscript{\rm 2}, 
    Qi Wu\textsuperscript{\rm 3}, 
    Si Liu\textsuperscript{\rm 1}\thanks{Corresponding author.}
    \\
}

\affiliations {
    \textsuperscript{\rm 1} Beihang University, Beijing, China \\
    \textsuperscript{\rm 2} Alibaba Group, Beijing, China \\
    \textsuperscript{\rm 3} University of Adelaide, Australia \\
    \{wzk1015, liusi\}@buaa.edu.cn, renda.brd@alibaba-inc.com, qi.wu01@adelaide.edu.au
}

\begin{document}

\maketitle

% \linenumbers

\begin{abstract}
When describing an image, reading text in the visual scene is crucial to understand the key information. Recent work explores the TextCaps task, \emph{i.e.} image captioning with reading Optical Character Recognition (OCR) tokens, which requires models to read text and cover them in generated captions. Existing approaches fail to generate accurate descriptions because of their (1) poor reading ability; (2) inability to choose the crucial words among all extracted OCR tokens; (3) repetition of words in predicted captions. To this end, we propose a Confidence-aware Non-repetitive Multimodal Transformers (CNMT) to tackle the above challenges. Our CNMT consists of a reading, a reasoning and a generation modules, in which Reading Module employs better OCR systems to enhance text reading ability and a confidence embedding to select the most noteworthy tokens. To address the issue of word redundancy in captions, our Generation Module includes a repetition mask to avoid predicting repeated word in captions. Our model outperforms state-of-the-art models on TextCaps dataset, improving from 81.0 to 93.0 in CIDEr. Our source code is publicly available \footnote{https://github.com/wzk1015/CNMT}.
\end{abstract}

\section*{Introduction}
\begin{figure}[t]
    \centering
    \includegraphics[width=0.47\textwidth]{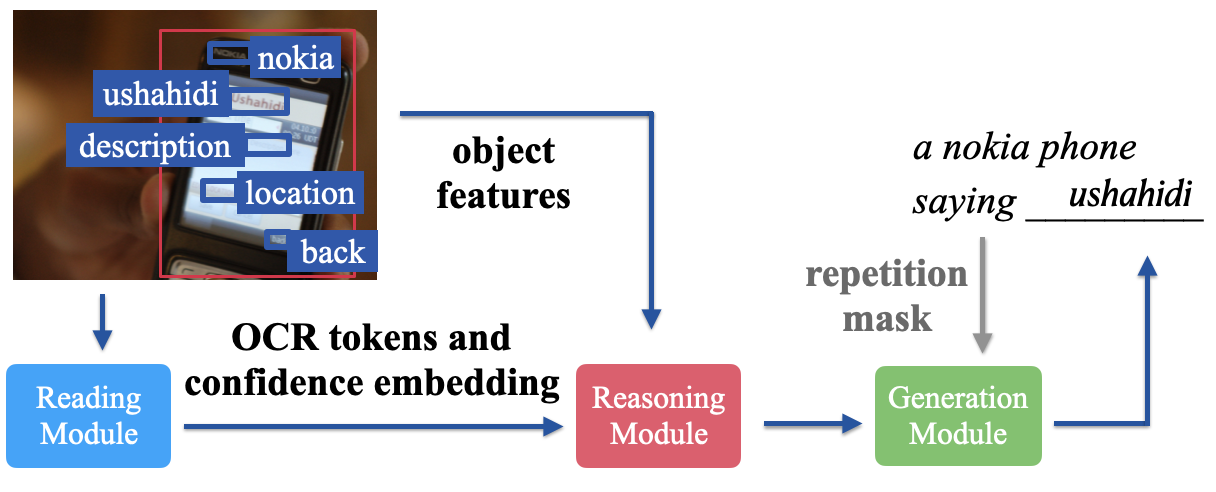}
    \caption{Our model extracts text in image with better OCR systems and records their recognition confidence as confidence embedding, which represents semantic importance of OCR tokens. After reasoning with objects and text features, it predicts caption tokens with a repetition mask to avoid redundancy.}
    \label{fig1}
\end{figure}

Image Captioning has emerged as a prominent area at the intersection of vision and language. However, current Image Captioning datasets \cite{coco,flickr} and models \cite{butd,aoanet} pay few attention to reading text in the image, which is crucial to scene understanding and its application, such as helping visually impaired people understand the surroundings. For example, in Figure \ref{fig1}, \emph{Ushahidi} on the screen tells the user the website he is browsing. To address this drawback, \citeauthor{textcaps} has introduced TextCaps \cite{textcaps} dataset, which requires including text in predicted captions.

In order to generate captions based on text from images, the model needs to (1) recognize text in the image with Optical Character Recognition (OCR) methods; (2) capture the relationship between OCR tokens and visual scenes; (3) predict caption tokens from fixed vocabulary and OCR tokens based on previous features. Current state-of-the-art model M4C-Captioner \cite{textcaps}, adapted to the TextCaps task from M4C \cite{m4c}, fuses visual modality and text modality by embedding them into a common semantic space and predicts captions with multi-word answer decoder based on features extracted from multimodal transformers \cite{transformer}.

While M4C-Captioner manages to reason over text in images, it is originally designed for TextVQA \cite{textvqa}, and thus fails to fit into Image Captioning task. It mainly has three problems. Firstly, its OCR system Rosetta \cite{rosetta} is not robust enough, making it suffer from bad recognition results. As words in captions come from either pre-defined vocabulary (common words) or OCR tokens, even a tiny error in recognizing uncommon words can lead to missing key information of the image.

Secondly, compared with answers in TextVQA where all OCR tokens can be queried in questions, captions in TextCaps should only focus on the most important OCR tokens in the image. In Figure \ref{fig1}, the key OCR tokens should be \emph{nokia} and \emph{ushahidi}, while others like \emph{location}, \emph{description} are dispensable. In fact, having these words in captions makes them verbose and even has negative effects. However, M4C-Captioner simply feeds all the OCR tokens into transformers without paying attention to their semantic significance, so irrelevant OCR tokens can appear in captions.

Thirdly, due to the use of Pointer Network \cite{pointernetwork} which directly copies input OCR tokens to output, M4C-Captioner's decoding module tends to predict the same word multiple times (\emph{e.g.} describe one object or OCR token repeatedly) in captions, like describing the image in Figure \ref{fig1} as \emph{a \textbf{nokia} phone saying \textbf{nokia}}. This redundancy leads to less natural captions which also misses key information \emph{ushahidi}, thus it should be avoided.

In this paper, we address these limitations with our new model Confidence-aware Non-repetitive Multimodal Transformers (CNMT), as shown in Figure \ref{fig1}. For the first issue, we employ CRAFT \cite {craft} and ABCNet \cite{abcnet} for text detection, and four-stage STR \cite{str} for text recognition. These new OCR systems help to improve reading ability of our model.

For the second issue, we record recognition confidence of each OCR token as a semantic feature, based on the intuition that OCR tokens with higher recognition confidence are likely to be crucial and should be included in captions, as they are frequently more conspicuous and recognizable. For instance, among all the OCR tokens in Figure \ref{fig1}, tokens with high recognition confidence (\emph{nokia}, \emph{ushahidi}) are consistent with our analysis on the key information in the image, while less recognizable ones (\emph{location}, \emph{description}) match dispensable words. Besides, tokens with lower confidence are more likely to have spelling mistakes. Therefore, we use recognition confidence to provide confidence embedding of OCR tokens. In Reasoning Module, OCR tokens and their recognition confidence are embedded together with multiple OCR token features, and fed into the multimodal transformers with object features of the image, fusing these two modalities in a common semantic space. 

For the third issue, we apply a repetition mask on the original pointer network \cite{pointernetwork} in the decoding step, and predict caption tokens iteratively. Repetition mask helps our model avoid repetition by masking out words that have appeared in previous time steps. We ensure that the repetition mask ignores common words such as \emph{a, an, the, of, says}, for they act as an auxiliary role in captions and are essential for fluency. As shown in Figure \ref{fig1}, at decoding step $t$, predicted score of \emph{nokia} is masked out as it appeared at step 2, allowing our model to generate the correct caption \emph{a \textbf{nokia} phone saying \textbf{ushahidi}} without repetition. Meanwhile, previously predicted common words \emph{a}, \emph{saying} is not affected in case of necessary repetition of them.

In summary, our contributions are threefold: (1) We propose our Confidence-aware Non-repetitive Multimodal Transformers (CNMT) model, which employs better OCR systems to improve reading ability, and uses confidence embedding of OCR tokens as representation of semantic significance to select the most important OCR tokens; (2) With the repetition mask, our model effectively avoids redundancy in predicted captions, and generates more natural captions; (3) Our model significantly outperforms current state-of-the-art model of TextCaps dataset by 12.0 in CIDEr on test set, improving from 81.0 to 93.0.

    \begin{figure*}[t]
    \centering
    \includegraphics[width=0.90\textwidth]{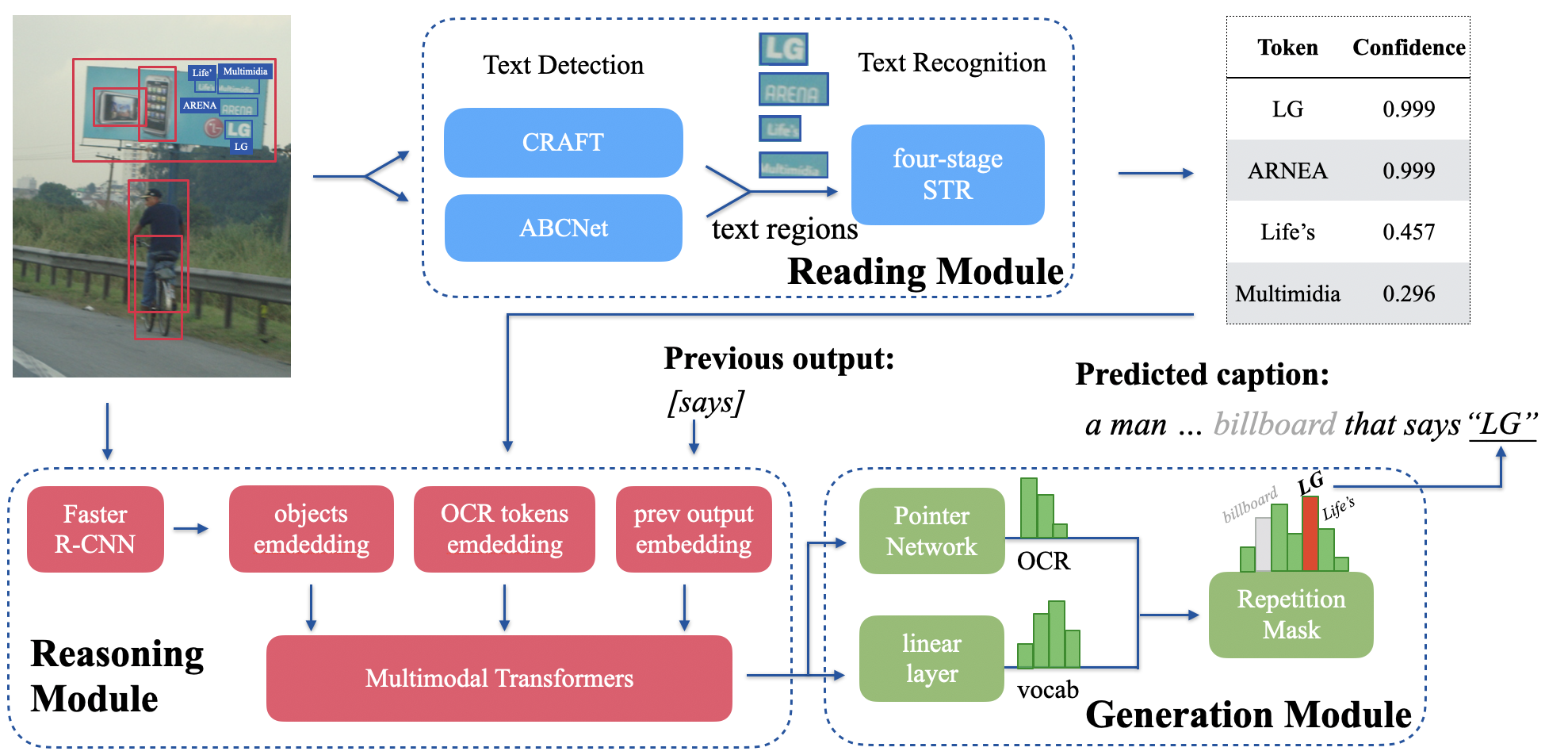}
    \caption{Overview of our CNMT model. In Reading Module, we extract OCR tokens with better OCR systems, and record their recognition confidence; Then Reasoning Module fuses OCR token features and object features with multimodal transformers, and Generation Module predicts caption tokens iteratively from a fixed vocabulary OCR tokens based on pointer network. A repetition mask is employed to avoid repetition in predictions.}
    \label{fig2}
    \end{figure*}

\section*{Related Work}

\smallskip
\textbf{Text based Image Captioning.}
In recent years, many works have focused on vision or language tasks \cite{reasoning, gao, liao}. Conventional Image Captioning datasets \cite{coco,flickr} aim to describe each image with a caption, but they tend to ignore text in the images as another modality, which is of great importance when describing the key information in the image. Recently TextCaps \cite{textcaps} dataset has been introduced, which requires reading text in the images. State-of-the-art models for conventional Image Captioning like BUTD \cite{butd}, AoANet \cite{aoanet} fail to describe text in TextCaps images. M4C-Captioner \cite{textcaps}, adapted from TextVQA \cite{textvqa} benchmark model M4C \cite{m4c}, is proposed to fuse text modality and image modality to make predictions. It employs multimodal transformers \cite{transformer} to encode image and text and predicts captions with an iterative decoding module. However, its performance is limited by poor reading ability and its inability to select the most semantically important OCR token in the image. Besides, its decoding module, originally designed for TextVQA task, shows redundancy in predicted captions. In this paper, we propose our CNMT model, which applies confidence embedding, better OCR systems and a repetition mask to address these limitations.

 \smallskip
\textbf{Optical Character Recognition (OCR).}
OCR helps to read text in images, which is crucial to TextCaps task. OCR involves two steps: detection (find text regions in the image) and recognition (extract characters from text regions). 
One way of text detection method is to use box regression adapted from popular object detectors \cite{textboxes}. Another method is based on segmentation \cite{textsnake}. 
For text detection, CRAFT \cite{craft} effectively detects text regions by exploring character-level affinity. Recent work ABCNet \cite{abcnet} presents a way to fit arbitrarily-shaped text by using Adaptive Bezier-Curve. For scene text recognition (STR), existent approaches have benefited from the combination of convolutional neural networks and recurrent neural networks \cite{crnn} and employment of transformation modules for text normalization such as thin-plate spline \cite{robust}. \citeauthor{str} introduce four-stage STR framework for text recognition. As for the TextCaps task, M4C-Captioner uses Rosetta \cite{rosetta} as OCR processor, but it is not robust enough to read text correctly. To solve this problem, our model adapts CRAFT \cite{craft} and ABCNet \cite{abcnet} as the detection module, and four-stage STR \cite{str} as the recognition module.

\section*{Methods}
\subsection*{Pipeline Overview}
Our CNMT is composed of three modules as shown in Figure \ref{fig2}. The input image is first fed into Reading Module to extract OCR tokens along with their recognition confidence, as the \emph{token-confidence} table on the top right part. Then Reasoning Module extracts object features of the image, embeds objects and OCR features into a common semantic space, and fuses them and previous output embedding with multimodal transformers. Finally, Generation Module uses output of the multimodal transformers and predicts caption tokens iteratively based on Pointer Network and the repetition mask, like predicting \emph{LG} at current step.

\subsection*{Reading Module}
As shown in the top part of Figure \ref{fig2}, Reading Module detects text regions in the image, and extract OCR tokens from these regions, jointly with confidence features of tokens.

\smallskip
\textbf{OCR systems.} 
We use two models for text detection, CRAFT \cite{craft} and ABCNet \cite{abcnet}. Text regions detected separately by CRAFT and ABCNet are combined together and fed into the text recognition part, as the four blue OCR boxes in the top part of Figure \ref{fig2}. For text recognition, we use deep text recognition benchmark based on four-stage STR framework \cite{str}. We combine OCR tokens extracted from our new OCR systems with the original Rosetta OCR tokens, and feed them into Reasoning Module.

\smallskip
\textbf{Confidence embedding.} 
Our previously mentioned intuition is that OCR tokens with higher recognition confidence tend to be crucial that should be included in captions, as they are frequently more conspicuous, recognizable and less likely to have spelling mistakes. Based on this, we record recognition confidence $x^{conf}$ of each OCR token from our text recognition system STR, where $x^{conf}$ is between 0 and 1. We then feed these confidence features into the next module to provide confidence embedding. As original Rosetta tokens do not include recognition confidence, OCR tokens that only appear in Rosetta recognition result are recorded by a default confidence value $c_{default}$. As shown in the top right part of Figure \ref{fig2}, we get several token-confidence pairs as the result of Reading Module.

\subsection*{Reasoning Module}
 For Reasoning Module we mainly follow the design of M4C-Captioner \cite{textcaps}, but with better OCR token embedding. As shown in the bottom left part of Figure \ref{fig2}, object features and OCR token features are jointly projected to a $d$-dimensional semantic space, and extracted by multimodal transformers.

 \smallskip
\textbf{Object embedding.} 
To get object embedding, we apply pretrained Faster R-CNN \cite{frcn} as the detector to extract appearance feature $x_m^{fr}$ of each object $m$. In order to reason over spatial information of each object, we denote its location feature by $x_m^b$ = $[x_{min}/W, y_{min}/H, x_{max}/W, y_{max}/H]$. The final object embedding ${x^{obj}_m}$ is projected to a $d$-dimensional vector as 
\begin{equation}
x^{obj}_m = LN(W_1x_m^{fr}) + LN(W_2x_m^b)
\end{equation}
, where $W_1$ and $W_2$ are learnable parameters, and $LN$ denotes layer normalization.

\smallskip
\textbf{OCR token embedding.} 
To get rich representation of OCR tokens, we use FastText \cite{fasttext}, Faster R-CNN, PHOC \cite{phoc} to extract sub-word feature $x^{ft}$, appearance feature $x^{fr}$ and character-level feature $x^{p}$ respectively. Location feature is represented as $x_i^b=[x_{min}/W, y_{min}/H, x_{max}/W, y_{max}/H]$. Then we add the confidence feature $x^{conf}$, based on the intuition that our model should focus more on tokens with higher recognition confidence. The final OCR token embedding ${x^{ocr}_i}$ is a list of $d$-dimensional vectors

\begin{equation}
\begin{aligned}
x^{ocr}_i = LN(W_3x_i^{ft} + W_4x_i^{fr} + W_5x_i^{p})\\
+ LN(W_6x_i^b) + LN(W_7x_i^{conf})
\end{aligned}
\end{equation}

where $W_3$, $W_4$, $W_5$, $W_6$ and $W_7$ are learnable parameters, and $LN$ denotes layer normalization.

\smallskip
\textbf{Multimodal transformers.} 
After extracting object embedding and OCR token embedding, a stack of transformers \cite{transformer} are applied to these two input modalities, allowing each entity to attend to other entities from the same modality or the other one. Decoding output of previous step is also embedded and fed into the transformers, like previous output \emph{says} in Figure \ref{fig2}. Previous decoding output $x^{dec}_{t-1}$ is the corresponding weight of the linear layer in Generation Module (if previous output is from vocabulary), or OCR token embedding ${x^{ocr}_n}$ (if from OCR tokens). The multimodal transformers provide a list of feature vectors as output:

\begin{equation}
    [z^{obj}, z^{ocr}, z^{dec}_{t-1}] = mmt([x^{obj}, x^{ocr}, x^{dec}_{t-1}])
\end{equation}

where $mmt$ denotes multimodal transformers.

\subsection*{Generation Module}

Generation Module takes output of multimodal transformers in Reading Module as input, predicts scores of each OCR token and vocabulary word, employs the repetition mask, and selects the predicted word of each time step, as shown in the bottom right part of Figure \ref{fig2}.

\smallskip
\textbf{Predicting scores.} 
Each token in the predicted caption may come from fixed vocabulary words $\{w^{voc}_n\}$ or OCR tokens $\{w^{ocr}_i\}$. Following the design of M4C-Captioner, we compute scores of these two sources based on transformer output $z^{dec}_{t-1}$ (corresponding to input $x^{dec}_{t-1}$). Scores of fixed vocabulary words and OCR tokens are calculated with a linear layer and Pointer Network \cite{pointernetwork} respectively. Pointer Network helps to copy the input OCR token to output. Linear layer and Pointer Network generate a $N$ dimensional OCR score $y_t^{ocr}$ and a $V$ dimensional vocabulary score $y_t^{voc}$. Here $V$ is the number of words in the fixed vocabulary and $N$ is pre-defined max number of OCR tokens in an image. This process can be shown as:

\begin{equation}
    {y_t^{ocr}} = PN(z^{dec}_{t-1}) 
\end{equation}
\begin{equation}
    {y_t^{voc}} = Wz^{dec}_{t-1} + b
\end{equation}
\smallskip
where $PN$ denotes Pointer Network. $W$ and $b$ are learnable parameters. 

Previous approaches consider scores of OCR tokens and vocabulary separately even if one word appears in both sources. However, this may lead to two sources competing with each other and predicting another inappropriate word. Therefore, we add scores of one word from multiple sources together to avoid competition. Adding scores of $n$-th vocabulary word can be described as:

\begin{equation}
y_{t,n}^{add} = y_{t,n}^{voc} + \begin{matrix} \sum_{i:w^{ocr}_i=w^{voc}_n} \ y_{t,i}^{ocr} \end{matrix}
\end{equation}

Then the final scores are the concatenation of added vocabulary scores and OCR scores:
\begin{equation}
y_{t} = [y_{t}^{add}, y_{t}^{ocr}]
\end{equation}

\begin{table*}[t]
    \centering
    \begin{tabular}{llllllllll}
    
    \hline\noalign{\smallskip}
    \# & Method & BLEU-4 & METEOR & ROUGE\_L & SPICE & CIDEr\\
    \noalign{\smallskip}\hline\noalign{\smallskip}
    1 & BUTD \cite{butd} & 20.1 & 17.8 & 42.9 & 11.7 & 41.9\\
    2 & AoANet \cite{aoanet} & 20.4 & 18.9 & 42.9 & 13.2 & 42.7\\
    3 & M4C-Captioner \cite{textcaps} & 23.3 & 22.0 & 46.2 & 15.6 & 89.6\\
    \noalign{\smallskip}\hline\noalign{\smallskip}
    \textbf{4} & \textbf{CNMT (ours)} & \textbf{24.8} & \textbf{23.0} & \textbf{47.1} & \textbf{16.3} & \textbf{101.7}\\
    \noalign{\smallskip}\hline
    
    \end{tabular}
    \caption{Evaluation on TextCaps validation set. We provide a comparison with prior works. Benefiting from better OCR systems, recognition confidence embedding and the repetition mask, our model outperforms state-of-the-art approach by a significant amount.}
    \label{table1}
    \end{table*}
    
    \begin{table*}[t]
    \centering
    
    \begin{tabular}{llllllllll}
    \hline\noalign{\smallskip}
    \# & Method & BLEU-4 & METEOR & ROUGE\_L & SPICE & CIDEr\\
    \noalign{\smallskip}\hline\noalign{\smallskip}
    1 & BUTD \cite{butd}  & 14.9 & 15.2 & 39.9 & 8.8 & 33.8\\
    2 & AoANet \cite{aoanet} & 15.9 & 16.6 & 40.4 & 10.5 & 34.6\\
    3 & M4C-Captioner \cite{textcaps} & 18.9 & 19.8 & 43.2 & 12.8 & 81.0\\
    \noalign{\smallskip}\hline\noalign{\smallskip}
    \textbf{4} & \textbf{CNMT (ours)} & \textbf{20.0} & \textbf{20.8} & \textbf{44.4} & \textbf{13.4} & \textbf{93.0}\\
    \noalign{\smallskip}\hline\noalign{\smallskip}
    5 & Human \cite{textcaps} & 24.4 & 26.1 & 47.0 & 18.8 & 125.5\\
    \noalign{\smallskip}\hline
    
    \end{tabular}
    \caption{Evaluation on TextCaps test set. Our model achieves state-of-the-art performance on all of the TextCaps metrics, narrowing the gap between models and human performance. }
    \label{table2}
    \end{table*}

\textbf{Repetition mask.} 
As we have mentioned in Section 1, repetition in captions brings negative effects on their fluency. In order to avoid repetition, we apply a repetition mask in Generation Module. At step $t$ of inference, the $N+V$ dimensional concatenated scores $y_t$ is added by a mask vector $M_t \in R^{N+V}$ , where the $i$-th element of $M_t$ is 
\begin{equation}
M_{t,i}=\left\{
\begin{array}{ccl}
-\infty & {if\ word_i\ appeared\ in\ previous\ steps}\\
0 & {otherwise}\\
\end{array} \right.
\end{equation}

$m$ is set to a minimum value. This helps to minimize the scores of elements that have appeared in previous steps, like the masked word \emph{billboard} in Figure \ref{fig2}.

Note that $M$ is applied only during inference. It focuses on repeating words, so when one word appears in both fixed vocabulary and OCR tokens or in multiple OCR tokens, all the sources will be masked out together. In addition, we ignore common words when applying mask, considering words like \emph{a, an, of, says, on} are indispensable to the the fluency of captions. Common words are defined as top-$C$ frequency words in ground-truth captions of training set, where C is a hyper-parameter.

\begin{figure}[h]
    \centering
    \includegraphics[width=0.44\textwidth]{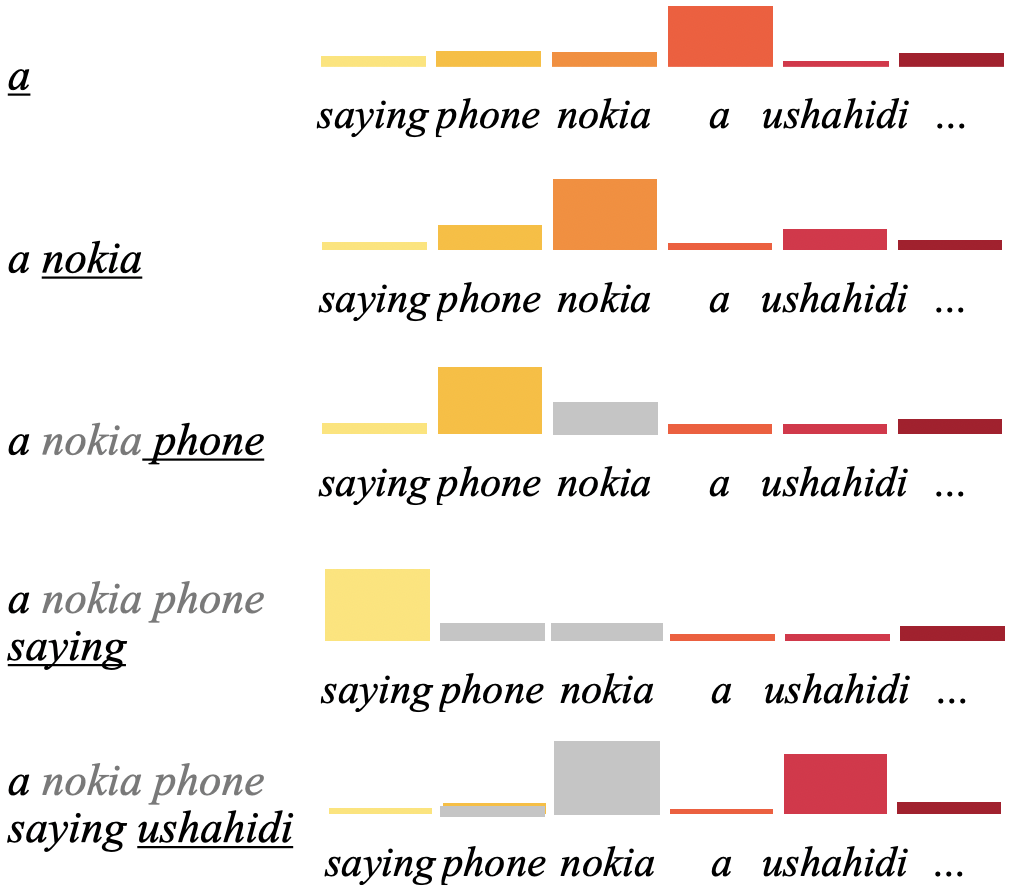}
    \caption{Illustration of the repetition mask. We show scores of words and predicted word at each step. \textcolor[RGB]{160,160,160}{Grey} indicates masked word. Common words like \emph{a}, \emph{saying} are ignored for their essentiality.}
    \label{fig3} 
    \end{figure}

In Figure \ref{fig3} we show an illustration of the repetition mask. Each row shows outputs(left) and predicted scores(right) at each decoding step. Since \emph{nokia} is predicted at step 2, its score is masked out from step 3 to the end (marked as grey). Scores of \emph{phone} are masked out from step 4. Common words \emph{a} and \emph{saying} are not masked. This mask prevents our model from predicting \emph{nokia} at the last step.

Therefore, the output word at step $t$ is calculated as
\begin{equation}
output_t=argmax(y_t+M_t)
\end{equation}

Our model iteratively  predicts caption tokens through greedy search, starting with begin token $\langle s \rangle$. Decoding ends when $\langle \backslash s \rangle$ is predicted.

\section*{Experiments}
We train our model on TextCaps dataset, and evaluate its performance on validation set and test set. Our model outperforms previous work by a significant margin. We also provide ablation study results and qualitative analysis.

\subsection*{Implementation Details}

For text detection, we use pretrained CRAFT \cite{craft} model and ABCNet \cite{abcnet} model with 0.7 confidence threshold. Affine transformation is applied to adjust irregular quadrilateral text regions to rectangular bounding box. We use pretrained four-stage STR framework \cite{str} for text recognition. For OCR tokens that only appear in Rosetta results, we set default confidence $c_{default} = 0.90$. We set the max OCR number $N=50$, and apply zero padding to align to the maximum number. The dimension of the common  semantic space is $d=768$. Generation Module uses 4 layers of transformers with 12 attention heads. The other hyper-parameters are the same with BERT-BASE \cite{bert}. The maximum number of decoding steps is set to 30. Words that appear $\geq$ 10 times in training set ground-truth captions are collected as the fixed vocabulary, together with $\langle pad \rangle$, $\langle s \rangle$ and $\langle \backslash s \rangle$ tokens. The total vocabulary size $V=6736$. Common word ignoring threshold $C$ of the repetition mask is set to 20.

The model is trained on the TextCaps dataset for 12000 iterations. The initial learning rate is 1e-4. We multiply learning rate by 0.1 at 5000 and 7000 iterations separately. At every 500 iterations we compute the BLEU-4 metric on validation set, and select the best model based on all of them. The entire training takes approximately 12 hours on 4 RTX 2080 Ti GPUs. All of our experimental results are generated by TextCaps online platform submissions.

\subsection*{Comparison with SoTA}
We measure our model's performance on TextCaps dataset using BLEU \cite{bleu}, METEOR \cite{meteor}, ROUGE\_L \cite{rouge}, SPICE \cite{spice} and CIDEr \cite{cider}, and mainly focus on CIDEr when comparing different methods, following the original TextCaps paper \cite{textcaps}.

We evaluate our model on TextCaps validation set and test set, and compare our results with TextCaps baseline models BUTD \cite{butd}, AoANet \cite{aoanet} and state-of-the-art model M4C-captioner\cite{textcaps}, as shown in Table \ref{table1} and Table \ref{table2}. Our proposed model outperforms state-of-the-art models on all five metrics by a large margin, improving by around 12 CIDEr scores on both validation set and test set. While the original gap between human performance and M4C-Captioner is 44.5 in CIDEr, our model narrows this gap by 27\% relative.

\subsection*{Ablation Study}
We conduct ablation study on OCR systems, confidence embedding and the repetition mask on validation set, and prove their effectiveness.

\smallskip
\textbf{Ablation on OCR systems.}
We first examine our new OCR systems through ablation study. We extract new OCR tokens with CRAFT and ABCNet and use four-stage STR for recognition, combine them with the original Rosetta OCR tokens, and extract their sub-word, character, appearance and location features. To focus on OCR system improvement, other parts of the model are kept consistent with M4C-Captioner. The result is shown in Table \ref{table3}. Compared with only using Rosetta-en, the model improves by around 3 CIDEr scores after employing CRAFT, and another 3 CIDEr scores after jointly employing ABCNet and CRAFT.

Another analysis can be seen in Table \ref{table4}, where we compute number of all OCR tokens and tokens that appear in ground truth captions to evaluate our OCR system improvement. After employing new OCR systems, total OCR tokens nearly tripled, and tokens that appear in ground truth captions nearly doubled, indicating our model's stronger reading ability. Jointly analyzing Table \ref{table3} and Table \ref{table4}, we conclude that better OCR systems lead to a larger amount of OCR tokens and thus higher probability to predict the correct word.

\begin{table}[t]
    \centering
    \begin{tabular}{llllllllll}
    \hline\noalign{\smallskip}
    OCR system(s) & CIDEr & \\
    \hline\noalign{\smallskip}
    Rosetta & 89.6 & - \\
    Rosetta + CRAFT  & 92.7 & (+3.1)\\
    \textbf{Rosetta + CRAFT + ABCNet} & \textbf{95.5} & \textbf{(+5.9)}\\
    \noalign{\smallskip}\hline
    
    \end{tabular}
    \caption{OCR systems experiment on TextCaps validation set. We keep other parts of the same configuration as M4C-Captioner in order to focus on OCR improvements. Our two detection modules CRAFT and ABCNet both bring significant improvements.}
    \label{table3}
    \end{table}

\begin{table}[t]
    \centering
    \begin{tabular}{lllllll}
    
    \hline\noalign{\smallskip}
    OCR system(s) & \# Total tokens  & \# In GT tokens\\
    \noalign{\smallskip}\hline\noalign{\smallskip}
     Rosetta-en             & 40.8k & 5.5k \\
    \noalign{\smallskip}\hline\noalign{\smallskip}
     Rosetta-en + &&\\
     CRAFT + ABCNet   & 117.5k & 10.0k \\
    \noalign{\smallskip}\hline\noalign{\smallskip}
    
    \end{tabular}
    \caption{OCR tokens analysis on validation set. We compare the original OCR system with our new ones, and demonstrate that both number of total OCR tokens and number of tokens that appear in ground truth captions have increased by a large amount.}
    \label{table4}
    \end{table}

\begin{table}[t]
    \centering
    \begin{tabular}{llllllllll}
    \hline\noalign{\smallskip}
    Method  & CIDEr & \\
    \hline\noalign{\smallskip}
    CNMT (w/o confidence) & 99.7 & -\\
    CNMT (multiply confidence) & 98.9 & (-0.8)\\
    \textbf{CNMT (confidence embedding)} & \textbf{101.7} & \textbf{(+2.0)}\\
    \noalign{\smallskip}\hline
    
    \end{tabular}
    \caption{Ablation of confidence embedding on validation set. Confidence embedding brings improvement on performance, while simply multiplying confidence to OCR token embedding leads to negative results.}
    \label{table5}
    \end{table}

    \begin{table}[t]
    \centering
    \begin{tabular}{llllllllll}
    \hline\noalign{\smallskip}
    Method  & Ignoring threshold $C$ & CIDEr\\
    \hline\noalign{\smallskip}
    CNMT (w/o mask) & - & 98.1\\
    CNMT & 0 & 92.6\\
    CNMT & 10 & 101.6\\
    \textbf{CNMT} & \textbf{20} & \textbf{101.7}\\
    CNMT & 50 & 99.4\\
    \noalign{\smallskip}\hline
    
    \end{tabular}
    \caption{Ablation of the repetition mask on validation set. Repetition mask helps to improve performance significantly. Experiment on hyper-parameter $C$ indicates that a small ignoring threshold has negative effects because of the essential auxiliary effects of these common words, while a large threshold limits the scope of the repetition mask.}
    \label{table6}
    \end{table}

\smallskip
\textbf{Ablations on confidence embedding.}
We evaluate the performance of OCR confidence embedding by ablating recognition confidence, as shown in Table \ref{table5}. Comparing line 1 and 3, we find that confidence embedding helps to improve performance by around 2.0 in CIDEr. This validates our intuition that recognition confidence serves as a way to understand semantic significance of OCR tokens and select the most important one when generating captions.

We compare our embedding method with a rather simple one: simply multiply recognition confidence (scalar between 0 and 1) to the final OCR token embedding ${x^{ocr}_i}$. Through this way, an OCR token is nearly a padding token (all zeros) if its confidence is small. However, as shown in line 2, this method actually brings negative effects, because it disturbs the original rich OCR token embedding. It also lacks learnable parameters, so the model is unable to decide the importance of confidence on its own.

\begin{figure*}[h]
    \centering
    \includegraphics[width=0.83\textwidth]{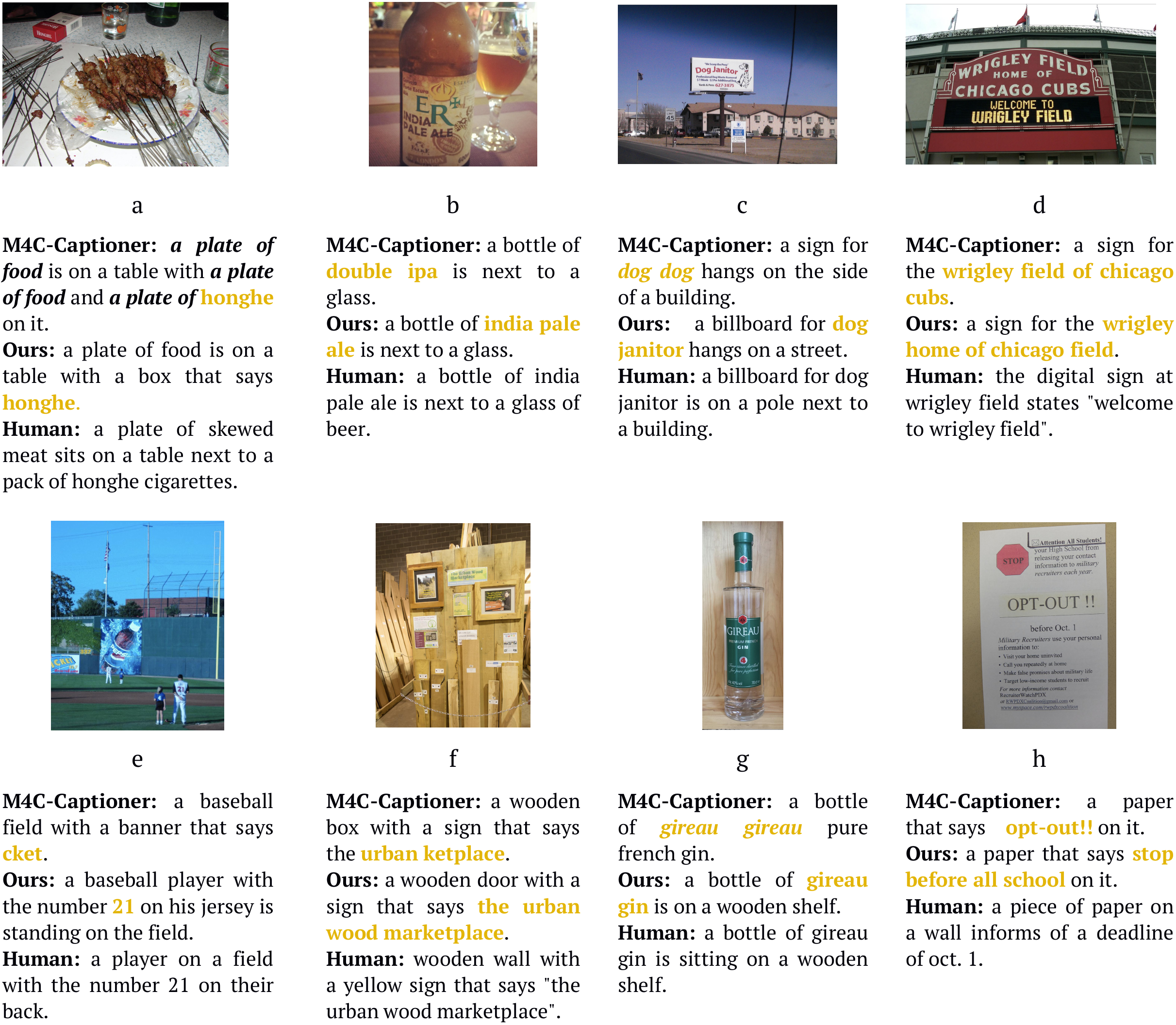}
    \caption{Qualitative examples on TextCaps validation set. \textcolor[RGB]{227,181,9}{Yellow} indicates words from OCR tokens. \emph{\textbf{Italic}} font indicates repetitive words. Compared to with previous work, our model has better reading ability, and can select the most important words from OCR tokens with confidence embedding. It also avoids repetition in predictions compared with M4C-Captioner.}
    \label{fig4} 
    \end{figure*}

\smallskip
\textbf{Ablations on the repetition mask.}
In Table \ref{table6} we provide ablation study on the repetition mask. It can be seen that the repetition mask improve performance by a relatively large amount of 3.6 in CIDEr. This proves our model's ability to predict more fluent and natural captions after removing repeating words, which solves an existing problem of previous approaches. Qualitative examples of therepetition mask can be found in Figure \ref{fig4} (a,c,g) where we give predictions of M4C-Captioner and our CNMT model, and prove our model's ability to avoid repetition effectively.

To prove the essentiality of ignoring common words when applying the repetition mask, we evaluate our model with an indiscriminate repetition mask, \emph{i.e.} all words include words like \emph{a}, \emph{an}, \emph{says} are masked out once they appear in previous steps. The result is shown in line 2 of Table \ref{table6}, where we find a large decrease in CIDEr, demonstrating the importance of ignoring common words. In fact, we find indiscriminate mask often generating unnatural captions, such as \emph{a poster for movie called kaboom with man holding gun} where articles \emph{a} are masked out, or \emph{a coin from 1944 next to other monedas} where \emph{money} is replaced with rarely used synonym \emph{monedas}. Such examples indicate that it is necessary to allow repetition of common words.

We conduct further experiments on hyper-parameter $C$, which is shown in Table \ref{table6}. When $C$ is set to a relatively small value, the repetition mask is applied on more commonly appeared words, and becomes indiscriminate when $C=0$. On the contrary, when $C$ is set to a large value, the scope of the repetition mask is limited, which brings negative effects. We observe that the best performance is achieved when $C$ is set to $20$.

\subsection*{Qualitative Analysis}
In Figure \ref{fig4} we provide example images of validation set and predictions from our model and M4C-Captioner. In Figure \ref{fig4} (e), with the help of confidence embedding, our model chooses the most recognizable OCR token \emph{21} instead of out-of-region word \emph{CKET} which is predicted by M4C-Captioner. Figure \ref{fig4} (b,f) shows our model's robust reading ability towards curved text and unusual font text. From Figure \ref{fig4} (a,c,g) we can see that our model significantly avoids repetition of words from both vocabulary and OCR tokens, and generates more fluent captions. While our model can detect multiple OCR tokens in the image, it is not robust enough to combine these tokens correctly, as shown in Figure \ref{fig4} (d) where our model puts the token \emph{field} in a wrong position. In Figure \ref{fig4} (h), our model fails to infer \emph{deadline} from text \emph{before Oct. 1}. As it requires more than simply reading, reasoning based on text remains a tough issue of predicting captions on the TextCaps dataset.

\section*{Conclusion}
In this paper we introduce CNMT, a novel model for TextCaps task. It consists of three modules: Reading module which extracts text and recognition confidence, Reasoning Module which fuses object features with OCR token features, and Generation Module which predicts captions based on output of Reading module. With recognition confidence embedding of OCR tokens and better OCR systems, our model has stronger reading ability compared with previous models. We also employ a repetition mask to avoid redundancy in predicted captions. Experiments suggest that our model significantly outperforms current state-of-the-art model of TextCaps dataset by a large margin. We also present a qualitative analysis of our model. Further research on avoiding repetition may include making the model learn by itself with reinforcement learning approach. As for semantic significance of OCR tokens, other features besides recognition confidence can be explored. We leave these as future work.

\section*{Acknowledgments}

This work was partially supported by the National Natural Science Foundation of China (Grant 61876177), Beijing Natural Science Foundation (Grant 4202034), Fundamental Research Funds for the Central Universities and Zhejiang Lab (No. 2019KD0AB04).

\bibliography{reference}

\end{document}